# An Agent based Approach towards Metadata Extraction, Modelling and Information Retrieval over the Web

Zeeshan Ahmed, Detlef Gerhard

Mechanical Engineering Informatics and Virtual Product Development Division (MIVP),
Vienna University of Technology,
Getreidemarkt 9/307 1060 Wien Austria
{zeeshan.ahmed, detlef.gerhard}@tuwien.ac.at

Web development is a challenging research area for its creativity and complexity. The existing raised key challenge in web technology technologic development is the presentation of data in machine read and process able format to take advantage in knowledge based information extraction and maintenance [4]. Currently it is not possible to search and extract optimized results using full text queries because there is no such mechanism exists which can fully extract the semantic from full text queries and then look for particular knowledge based information.

Mechanism of presenting information over the web in a format so that the humans as well as machines can understand the context leads to the concept of Semantic Web introduced by Tim Berners Lee [4]. Semantic web is a linked mesh of information to produce technologies capable of reasoning on semi structured information and processed by machines [4]. Although a number of semantic based solutions have been developed including Semantic Desktop [2], Reisewissen [1] and Metadata Search Layer [3] but still the web based problems are not completely solved.

We propose an approach Semantic Oriented Agent based Search (SOAS) to provide comprehensive support in implementing a semantic and agent based solution towards the problems of Meta data extraction, modeling and information retrieval over the web. SOAS is dynamic components and agents based approach to handle user's unstructured full text requests by converting in to structured information models and generating semantic based queries to search particular information. SOAS is designed by keeping four major requirements in mind, which are Automatic user request handling, Dynamic unstructured full text data read, analysis and modeling, Semantic query generation and Optimized result classifier.

The proposed designed architecture of SOAS, as shown in Fig. 1 is consists of one Personal Agent (PA) and five dynamic processing units .i.e., Request Processing Unit (RPU), Agent Locator (AL), Agent Communicator (AC), List Builder (LB) and Result Generator (RG) and Database. The architecture is designed in a way that the unstructured full text user request desiring some particular knowledge based information is given to PA. PA will simply transfer the received request to RPU, which will analyze and convert unstructured full text user request in to a structured semantic based data request. Constructed new semantic based search request will be passed and used by AL to find out the contact information of particular domain's available agents. Then using extracted contacts AC will contact, communicate, obtain information from available agents and stored in to the database. Stored information will be extracted and prioritized list of results will be generated by LB. Prioritized results will be forwarded to RG which will finalize the result by converting into the acceptable readable format and then will pass to PA. In the end PA will respond back to user by providing retrieved results.

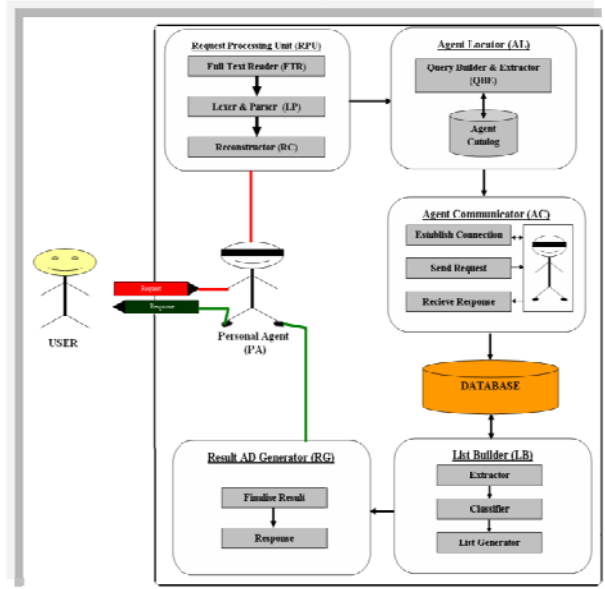

**Fig. 1.** The architecture of Semantic Oriented Agent based Search (SOAS)

We have shortly illustrated that SOAS can be well suited for supporting Semantic Web by providing a comprehensive information extraction and modeling process. In future to evaluate the effectiveness of SOAS we will implement it in a real time software application using existing semantic web and desktop technologies.